\documentclass[12pt]{article}

\usepackage[english]{babel}
\usepackage{authblk}
\usepackage[a4paper,top=2.5cm,bottom=2.5cm,left=3cm,right=3cm,marginparwidth=2.75cm]{geometry}
\usepackage{marginnote}
\usepackage{amsmath,amssymb}
\usepackage[english]{babel}
\usepackage{amsthm}
\usepackage{graphicx}
\usepackage[colorlinks=true, allcolors=blue]{hyperref}
\usepackage{algorithm}
\usepackage{algpseudocode}
\usepackage{wrapfig}

\newtheorem*{conjecture}{Conjecture}
\newtheorem*{corollary}{Corollary}

\DeclareMathOperator*{\argmax}{argmax}
\DeclareMathOperator*{\argmin}{argmin}

\title{Permutative redundancy and uncertainty of the objective in deep learning}
\author{Vacslav Glukhov}
\affil{glukhov@stanfordalumni.org}
\begin{document}
\maketitle

\begin{abstract}
Implications of uncertain objective functions and permutative symmetry of traditional deep learning architectures are discussed. It is shown that traditional architectures are polluted by an astronomical number of equivalent global and local optima. Uncertainty of the objective makes local optima unattainable, and, as the size of the network grows, the global optimization landscape likely becomes a tangled web of valleys and ridges. Some remedies which reduce or eliminate ghost optima are discussed including forced pre-pruning, re-ordering, ortho-polynomial activations, and modular bio-inspired architectures.  
\end{abstract}

\section{Introduction}

The success of practical applications of deep networks trained with stochastic gradient descent and its variants is undisputed. Despite occasional progress, theoretical research on the foundations of deep networks is lagging behind their practical successes. Deep learning theories often rely on assumptions that are far from realistic, e.g., availability of infinite training data and access to statistical expectations, regularity -- determinism, continuity, smoothness -- of objective functions, and infinite production run times. Aleatoric uncertainty is often treated as a mere inconvenience which can be averaged away. Ignorance of epistemic uncertainty is common.  

This is unfortunate. Better theoretical foundations help guide practitioners to systematically reason about their models and consciously architect cognitive machines rather than stumble upon solutions through trial and error. 

It is evident that with sheer computing power, a deep enough and wide enough approximator can model almost any complex real-world phenomenon. The latest SoTAs often appear as competitions in tackling complex problems with brutal force rather than ingenuity and creativity. In all this excitement, have we forgotten that the point is not to compute more, but to create better - efficient, robust, safe, and transparent - solutions for 
real-world applications? 

Let us talk about efficiency. The brain is a super-efficient cognitive machine. It consumes only about 20 watts of energy, give or take a few watts. It does not pass a continuous flow of information through its internal pathways when these pathways are not needed. It does not learn by error backpropagation, nor does it adjust its connections using gradients. The brain does not optimize any global objectives in the sense we use the term optimization in deep learning. Nor is it a feed-forward machine. 

Deep learning \cite{DeepLearning_LeCun_2015}, for which feedforward monolithic architectures, learning by gradient optimization, and backpropagation are key, is a breakthrough in technology that opened many doors. At the same time, the regrettable side effect of this success is that it diverts talent and technical resources on a massive scale -- away from other research and development directions in computational cognition. 

We are reaching the point of exponentially diminishing returns when using brute computing of deep networks to solve complex real-world problems. 

I would like to draw attention to these architectures' fundamental limitations. Here, based on well-known but almost neglected theoretical and experimental results, I discuss the implications of gradient optimization of uncertain objectives and permutative redundancies of deep networks.

I also discuss some remedies. 

\section{Uncertain objectives in an uncertain world}

Gradient methods are commonly used to optimize deep networks. Architectures are chosen to efficiently evaluate the objective and its gradients in the parameter space via a forward pass of data inputs through the depth of the network. For some problems, the network's objective is to approximate dependent variables in the data or their distribution, given the set of independent or state variables. For other problems, e.g. in the reinforcement learning settings, the network approximates a strategy that links the state variables with actions, and the objective minimizes cumulative loss (or maximizes cumulative reward) incurred during the execution of said strategy. In almost all cases, excluding perhaps clearly degenerate ones, the values of the objective and its gradients are uncertain.

To handle aleatoric or statistical uncertainty, the known unknowns,  data-centric methods rely on the availability of massive amounts of data to handle: machine learning as a branch of applied statistics works only when there is plenty of data. Not every situation provides enough data to alleviate statistical uncertainties. Moreover, data are tenuous and statistical uncertainty stubbornly persists in the most interesting situations and problems.

An abundance of data as a requirement places natural limits on all data-centric methods.  

Consider a real-world agent trained to produce an optimal strategy with a variant of reinforcement learning. The agent must adapt to uncertain and changing environmental conditions. Three issues arise here. 
\begin{description}
    \item[First] When the rate of change exceeds the rate with which the agent updates its knowledge, learning an optimal strategy might be too expensive, and the agent's best attainable strategy inevitably remains suboptimal.
    \item[Second] When the agent learns about changing conditions only or mostly by examining the statistical properties of its rewards, it must always continue exploration to avoid being crushed by the evolving environment. This, again, leads to the local suboptimality of the agent's performance. 
    \item[Third] An agent architectured to survive epistemic uncertainty (i.e. unknown unknowns or unknowable unknowns\footnote{The following passage is attributed to Donald Rumsfield, the US Secretary of Defense: \textit{Reports that say that something hasn't happened are always interesting to me, because as we know, there are known knowns; there are things we know we know. We also know there are known unknowns; that is to say we know there are some things we do not know. But there are also unknown unknowns—the ones we don't know we don't know. And if one looks throughout the history of our country and other free countries, it is the latter category that tends to be the difficult ones.} } \footnote{The following passage is often quoted from Frank Knight's Risk, Uncertainty and Profit \cite{Knight1921}:\textit{ ...essential fact is that 'risk' means in some cases a quantity susceptible of measurement, while at other times it is something distinctly not of this character; and there are far-reaching and crucial differences in the bearings of the phenomena depending on which of the two is really present and operating}}) must maintain broadly tuned cognitive machine capable of recognizing not only statistically measurable regime shifts but also react reasonably in response to unmeasurable environmental changes. 
\end{description}

These very practical issues of the uncertainty of objectives or environment cannot be answered by data-centric methods alone -- the answers lie beyond data.
    
\centerline{***}

It would be a mistake to imply that deep learning practitioners ignore uncertainty. On the contrary, in all practical applications, we use batches and take advantage of symmetries in the probability distribution function of the objective to reduce uncertainty. And it works - to an extent. In many cases, the observed convergence of the gradient methods, and demonstrable practical usefulness of the trained models typically attest that training achieves its goals. In other cases, however, tedious tweaking and tinkering with gradient methods is necessary to achieve convergence at the cost of massive computational expense and enormous data requirements. But we do not seem to have a systematic view of why what we do works for some problems and is not for others.  

In the following sections, I discuss how the real-world aspect of deep learning, the uncertainty of objectives, conflicts with the core architectural characteristics of deep networks --  the permutability of their elements. I explore some remedies. 

Elsewhere I aim to discuss how the desideratum of adaptation to the uncertain environment of real world agents also clashes with traditional deep learning architectures.

\section{Representation of uncertain objective functions}

Given the data sets representing the inputs $x_i\in X$ and outputs $y_i\in Y$ the network approximates $Y = G(\theta, X)$. Optimization is looking for an extremum (a minimum for definitiveness) of a loss function:
\begin{equation}
    \theta^* = \argmin_\theta \mathcal{L}(Y,G(\theta,X))
\end{equation}
For a regression or classification problem, the loss function may look like
\begin{equation}
    \mathcal{L}(Y,G(\theta,X)) =  \| Y - G(\theta,X) \| \label{eq:lossL}
\end{equation}
where the norm $\|\cdot\|$ is suitably chosen for each particular case. A quadratic norm is almost always a default. 

\subsection{Note on the structure of the optimization objective}

The choice of the norm in Eq \ref{eq:lossL}, although routine, is by no means self-evident or often even compelling.  Remember that $\mathcal{L}$ is assumed to represent an approximation of the real-world losses associated with imperfect modelling and risk. In practice, losses often are much less convenient than a quadratic norm is capable of representing: they have a non-trivial structure, are often asymmetric, statistical estimates in Eq \ref{eq:lossL} can misbehave and have no well-defined or reasonably attainable limit when normalized by the number of data points. Researchers adjacent to machine learning areas of applied statistics have long realized that a realistic structure of losses due to an imperfect model and uncertainty is crucially important for mission-critical applications (see for example \cite{Viole2017} and \cite{Nawrocki:Viole:RePEc:eee:finana:v:95:y:2024:i:pb:s1057521924003752} and references therein). I will discuss these issues elsewhere in more detail.

\subsection{Quadractic structure}
For our purposes, it makes sense to simplify the notation and adopt the objective function of data and parameters: 
\[
    \theta^*(D) = \argmin_\theta  \mathcal{J}(\theta, D) 
\]
where $D $ denotes data. The notation is general enough to describe a sample of $\mathcal{J}$, as well as its statistical averages. The explicit dependency of the optimal set of parameters $\theta^*(D)$ on data $D$ is important.

The idealized representation of the "true" objective function of parameters often used in studying the mathematical foundations of deep learning is 
\begin{equation}
    \mathcal{J}(\theta, D) = (\theta - \theta^*(D))^\prime H(D) (\theta - \theta^*(D)) \label{eq:J-representation}
\end{equation}
where $H$ is the Hessian matrix. The assumption of the existence of a regular $\mathcal{J}(\theta)$ that does not depend on data in the limit of an infinite data set is often made in the literature. 
\[
\mathcal{J}(\theta) = \lim_{\|D\|\rightarrow \infty} \mathcal{J}(\theta, D) 
\]
It is a strong assumption: the rest immediately falls in place once you accept a smooth, twice differentiable structure of $\mathcal{J}(\theta)$. It is also clearly a flawed assumption: data are always limited.

There are many ways to represent uncertainty in the objective function. For example, in \cite{pmlr-v97-ghorbani19b} the following representation is adopted: 
\begin{equation}
    \mathcal{J}(\theta,D) = (\theta - \theta^*(D) + \epsilon)^\prime H(D) (\theta - \theta^*(D) + \epsilon)  \label{eq:J-representation-noise-pmr}
\end{equation}
This form is also compatible with that used in \cite{SGD_Yuan_Ying_Vlaski_Sayed} where the authors studied the effects of sampling noise. An implicit assumption here is that the magnitude of uncertainty is a function of the Hessian.

I am interested in making as few implicit assumptions as possible. To better capture the real-world properties of the objective function I adopt the following representation which differs slightly from Eq \ref{eq:J-representation-noise-pmr}:
\begin{equation}
    \mathcal{J}(\theta, D) = (\theta - \theta^*(D))^\prime H(D) (\theta - \theta^*(D)) + \hat{\epsilon}(\theta, D) \label{eq:J-representation-noise}
\end{equation}
Representation in Eq \ref{eq:J-representation-noise} explicitly accounts for the fact that the optimum of $\mathcal{J}(\theta^*)$ is a function of data. $D$ denotes the data set available for training, testing, or running, $\theta^*(D)$ and $H(D)$ indicate that the optimum set of parameters and the Hessian are generally functions of available data.  

Gradient iterates require knowledge of $\nabla_\theta \hat{\epsilon}(\theta,x)$. I assume that the uncertain gradient $\nabla_\theta \mathcal{J}(\theta,x)$ is available, which is the case for most modern network architectures, and $\epsilon(\theta, x)$ represents the stochastic part of the gradient, so that
\begin{equation}
    \nabla_\theta \mathcal{J}(\theta,D) + \epsilon(\theta,D) \label{eq:nablaj}
\end{equation}
The first part in Eq \ref{eq:nablaj} is deterministic subject to the data set $D$. In practice, we assume that gradients $\nabla_\theta$ are computed and evaluated automatically, and it is valid that, evaluated on the same set of inputs $D$, uncertainty is reasonably regular:
\begin{equation}
    \| \hat{\epsilon}(\theta^\prime,D) - \hat{\epsilon}(\theta,D) \| < L_{\epsilon}(\theta) \|\theta^\prime-\theta\| < \infty
\end{equation}
The magnitude of the gradient uncertainty is 
\begin{equation}
    \|\epsilon(\theta)\| \sim L_\epsilon(\theta) < \infty   
\end{equation}
%A more rigorous study will shed more light on the properties of the uncertain objective in the future -- if necessary. 
Many authors also make a strong assumption of a zero-expectation, finite-variation uncertainty:
\begin{equation}
    \mathbb{E}( \epsilon(\theta, D) ) = 0 \label{eq:norm_epsilon}
\end{equation}
\begin{equation}
    \forall D: \mathbb{E}(\epsilon^2(\theta, D)) < \infty. \label{eq:var_epsilon}
\end{equation}
Expectations are convenient statistical abstractions stemming from an assumption that uncertainty can theoretically always be averaged away, in the long, unlimited run, given enough samples in training, testing, and running.

Truth is all real-world deep learning models are {finite-sample and limited-run.}   Expectations $\mathbb{E}(\cdot)$ are a mathematical fiction. In the absence of long-run statistical quantities (probabilities, averages, etc) a practitioner must explicitly account for the finiteness of all samples to control all approximations.  A constructive rule-of-thumb is to keep in mind that "zero is dimensional quantity", meaning that when a statistical quantity such as the empirical average of noise is declared small one must always specify -- small compared to what?

Let me now proceed to the main discussion. 

\section{Gradient convergence near a presumed local quadratic optimum under an uncertain objective} \label{sec:grad_convergence}

Sampling noise is an obvious source of uncertainty of the objective.  Convergence of gradient methods for non-deterministic objective functions with sampling noise has been studied in  \cite{SGD_Yuan_Ying_Vlaski_Sayed} and references therein. Another general study of stochastic gradient convergence with noisy gradients is in \cite{dieuleveut2018bridginggapconstantstep}, see their references as well. 

For this work, the main result of the above studies is that contrary to the case of a deterministic objective function, where gradient methods converge exponentially fast, convergence toward a noisy objective does not necessarily hit an optimum. Instead, the gradient iterates eventually reach a stationary vicinity in the parameter space around the optimum. The shape and the size of the equilibrium vicinity depend on the learning rate, the extent of the noise, and the structure of the deterministic part of the objective function near the optimum. Authors of \cite{SGD_Yuan_Ying_Vlaski_Sayed} and \cite{dieuleveut2018bridginggapconstantstep} approach the problem differently but arrive at similar conclusions. 

Practitioners mostly ignore this distinction between deterministic (on average) and inherently stochastic systems, and assume that some clever reduction of the learning rate, the key hyper-parameter in all variants of gradient methods, should eventually drag the gradient iterates toward the optimum. What we know is that convergence to an optimum is slow, and, to my knowledge, there is neither general formal proof, nor guarantee of convergence. In the case of a purely quadratic objective function in the vicinity of the optimum, the only result that is reasonably well proven is that the expectation of the gradient iterates does approach the true optimum -- but only eventually, in the long run. Expectations, however, are not practical because training always deals with a limited data set and production runs require definite values of the model parameters. Even if we resort to using averages in place of expectations, it is unclear how to practically create a deep network configuration corresponding to an average set of parameters. We don't know and cannot presume that a model with "average" values of parameters will even work:
\begin{equation}
    \overline{\mathcal{J}(\theta)} \neq \mathcal{J}(\overline{\theta})
\end{equation}
Quadratic approximation might be a convenient theoretical tool, but, as we will see shortly, the true shape of the objective function in the vicinity of an optimum is far from a smooth quadratic form in Eqs \ref{eq:J-representation}, \ref{eq:J-representation-noise-pmr} or \ref{eq:J-representation-noise}.

\begin{center}
    ***
\end{center}
Following \cite{pmlr-v97-ghorbani19b} and \cite{sagun2017eigenvalueshessiandeeplearning} and \cite{dieuleveut2018bridginggapconstantstep}, I derive the expression for the variance of the model parameters in the case of an uncertain objective function $\mathcal{J(\theta,D)}$ where $\theta$ are the parameters, and $D$ is the input data. 

In practice, we use a portion of the training data set to estimate the gradient iterates. Let $B$ be a batch of data. Without loss of generality in the local vicinity of the optimal $\theta^*$, as in Eq \ref{eq:J-representation-noise}:
\[
\mathcal{J}_e(\theta, B) \sim\frac{1}{2} \left(\theta - \theta^*(D_T)\right)^\prime H (\theta - \theta^*(D_T))  + \hat{\epsilon}(\theta, B),
\]
where $\theta^*(D_T)$ is a hypothetical optimal value of parameters given the training data set and $\epsilon(\theta,B)$ is the uncertainty, or noise, of the objective function. As I already discussed, it is not unreasonable to assume that, evaluated on the same batch $B$, the actual sample in the close vicinity of $\theta$ is close to the sample at $\theta$  
\begin{equation}
\hat{\epsilon}(\theta + \delta \theta, B) = \hat{\epsilon}(\theta, B) + \epsilon(\theta, B)\cdot \delta \theta 
\end{equation}

Access to the uncertainty of the objective function's gradient is almost always available in practical deep learning applications. Gradient iterates are:
\begin{equation}
\theta_{t+1} = \theta_t - \eta\cdot H \cdot (\theta_t -\theta^*) + \eta \cdot \epsilon(\theta,B_t)\label{eq:Iterates}
\end{equation}
where $\eta$ is the learning rate. Eq \ref{eq:Iterates} is an AR(1) auto-regressive process.  

Unrolling the iterates for $\theta_t$ as in \cite{pmlr-v97-ghorbani19b} modulo the definition of $\epsilon$, we can formally write:
\begin{equation}
    \theta_t -\theta^*= (I - \eta \cdot H)^t (\theta_0 - \theta^*) - \eta \cdot \sum_{j=0}^{t-1} (I - \eta\cdot H)^j \epsilon(\theta_{t-j-1}, B_{t-j-1}) \label{eq:thetastart}
\end{equation}

The first term in the above Eq \ref{eq:thetastart} expresses the well-known result that gradient iterates forget the initial state exponentially fast. The second term is just the unrolled standard vector autoregression of the noise.

Let us now study the stochastic properties of the increments of the gradient iterates. Let $q_i$ be a unit vector along the $i$-th axis of an $n \times n$ diagonalized Hessian $H = \text{diag}(\lambda_1, \dots, \lambda_n)$ where $\lambda_1 > \lambda_2 > \dots \lambda_n$ Then

\begin{equation}
\begin{split}   
    q_i (\theta_{t+1} - \theta_t)  = &- \eta \lambda_i\left\{ (1 - \eta \lambda_i)^t\right\} q_i (\theta_0 - \theta^*)  \\
    &- \eta \left[ q_i \epsilon_t  + \sum_{j=1}^{t} (1-\eta \lambda_i)^j q_i\epsilon_{t-j}  - \sum_{j=0}^{t-1} (1-\eta \lambda_i)^jq_i\epsilon_{t-j-1}\right] \\
    = & \underbrace{- \eta \lambda_i\left\{ (1 - \eta \lambda_i)^t\right\} q_i (\theta_0 - \theta^*)}_{\rightarrow 0, \text{deterministic decay term}} \\
    &\underbrace{ -\eta \left[ q_i\epsilon_t - \eta \lambda_i \sum_{j=0}^{t-1} q_i\epsilon_{t-j-1} (1-\eta\lambda_i)^j\right]}_{\text{cumulative stochastic term}}
\end{split}\label{eq:det_stoch}
\end{equation}

Under the standard assumptions $0 < \eta \lambda_i < 1$ the deterministic term in Eq \ref{eq:det_stoch} eventually decays to 0. The stochastic term in Eq \ref{eq:det_stoch} accumulates uncertainty over approximately $\tau_i$ iterates, where
\begin{equation}
    \tau_i \sim \frac{1}{\eta \lambda_i}. \label{eq:tau_i}
\end{equation}
so that %\marginnote{No, revise. Yes, done.}
\begin{equation}
    \eta \lambda_i \sum_{j=0}^{t-1} q_i\epsilon_{t-j-1} (1-\eta\lambda_i)^j \approx \frac{1}{\tau_i} \sum_{j=1}^{\tau_i} \epsilon_{t-j} \label{eq:stoch_tau_i}
\end{equation}
and the magnitude of the stochastic term is essentially 
\begin{equation}
    \| q_i \delta \theta \|^2 \sim \eta^2 \| q_i \epsilon \|^2 \label{eq:vartheta}
\end{equation}

In theoretical work, the assumption of a finite variance of $\epsilon$ in Eq \ref{eq:vartheta} is often crucial. In practice, there is no guarantee that the distribution of the objective's uncertainty is regular. 

\subsection{Boltzmann's equilibrium around the presumed optimum}

The dynamics of gradient iterates in the presence of stochasticity is essentially a discrete version of the stochastic dynamics described by the overdamped Langevin equation, or, in a more general case, by a Fokker-Plank equation:
\begin{equation}
    \partial P(\theta,t) /\partial t = \left\{ \nabla^2\cdot \frac{\eta^2\cdot \|\epsilon\|^2}{2} - \eta \cdot \nabla \cdot  J(\theta) \right\} P(\theta, t) \label{eq:Fokker}
\end{equation}
In Eq \ref{eq:Fokker} both differential operators act on everything to their right following the standard chain rule.

It is a well known fact that under the assumption of a finite variance of the noise term, the equilibrium distribution of the continuous version of the overdamped Langevin equation reproduces a Boltzmann distribution (see the excellent treatment of the topic in \cite{SCH00x}).

\begin{equation}
    P(\theta, t\rightarrow \infty) \sim \exp \left(-\beta \cdot J(\theta) \right) \label{eq:boltzmann}
\end{equation}
where the temperature governing the extent of the distribution in the parameter space is
\begin{equation}
    \beta^{-1} \sim  \eta \|\epsilon \|^2  \label{eq:beta}
\end{equation}

To summarize, relaxation to the equilibrium distribution occurs exponentially fast first, governed by the deterministic part of the gradient, and then slowly, governed more by the stochastic term and the anisotropy of the Hessian. A clever choice of the learning rates (equivalently, freezing schedule) reflecting the anisotropy of the Hessian helps improve the convergence rate in situations of low uncertainty. See, for example, the description of Adam in \cite{kingma2017adammethodstochasticoptimization} and the classic text \cite{NIPS2012_c399862d}. Similar methods can indeed be applied in the situation of high uncertainty of the objective, but the optimum might prove unreachable within a reasonable computing time and with a reasonable amount of data. 

\subsection{Effects of the anisotropy of the Hessian}

In the preceding, I reproduce a well-known but often overlooked result is that, for a noisy objective function, the stochastic gradient method and its derivatives with a finite learning rate converge not to the optimum set of parameters but to an equilibrium distribution around the optimal point. The noise variance and the objective function's curvature near the optimum define the equilibrium distribution of the model parameters.

Then, from Eqs \ref{eq:boltzmann}  and \ref{eq:beta} the ball of a radius of at least 
\begin{equation}
    R_\theta^* \sim \sqrt{\frac{ \eta \|\epsilon\|^2}{\lambda_1}}
\end{equation}
in the parameter space surrounds the optimum $\theta^*$. The following property is going to be used shortly: the volume of the parameter space occupied by the gradient iterates is
\begin{equation}
    V_\theta^* \sim \frac{\pi^{n/2}}{\Gamma(n/2 +1)} (\eta \|\epsilon\|^2)^{1/2} \prod_{i=1}^n \lambda_i^{-1/2}\label{eq:vol_ball}
\end{equation}

Lastly, the magnitude of the stochastic term in the equation for the gradient iterates increments  \ref{eq:det_stoch} depends on the number of effective batches participating in the averaging along the trajectory of the gradient iterates. The effective number of batches in the direction of $i$ is defined by 
\[
\tau_i = -\frac{1}{\log(1-\eta \lambda_i)} \approx \frac{\lambda_1}{c\lambda_i}.
\]

This number is small for directions $i$ in the parameter space corresponding to the largest eigenvalues of the Hessian. The objective function uncertainty is well averaged for the flatter directions, where the convergence is slow, but is not so well averaged for the directions of greater curvatures. The sampling noise alone may explain the slow convergence of the stochastic gradient method and its derivatives for some real-world systems. All we know now is that sampling uncertainty is reduced in the variable rate gradient methods such as Adam \cite{kingma2017adammethodstochasticoptimization} but is never eliminated.

Let me briefly touch on the possible implications of anisotropy and of the Hessian and gradient iterates. Eqs \ref{eq:det_stoch}, \ref{eq:tau_i}, and \ref{eq:stoch_tau_i} also imply that the level of uncertainty may depend on the direction. We can approximate the equilibrium vicinity of gradient iterates around the optimum with 
\begin{equation}
    P(\theta) \sim  \exp \left( - \sum_i^n \beta_i \lambda_i (q_i\cdot(\theta - \theta^*))^2\right)
\end{equation}
and, if Eq \ref{eq:stoch_tau_i} holds,
\begin{equation}
    \beta_i^{-1} \sim \eta \cdot \lambda_i \cdot \| q_i\cdot\epsilon \|^2 
\end{equation}
which in turn indicates that the size of the equilibrium vicinity does not depend on the direction in the parameter space as the two anisotropies - the Hessian, and the uncertainty of the objective along the trajectory of the gradient iterates, cancel out. I will discuss this surprising result elsewhere. I will present some experimental results as well.

I will now demonstrate how the issues associated with the uncertainty of the objective affect deep learning. 

\section{Implications of uncertain objectives and permutative redundancy of deep networks}

Modern deep learning architectures rely on computational structures that consist of multiple layers or blocks of relatively simple interchangeable and permutable input-output elements. The ability of these structures to approximate the complexity of relationships found in the real world is the key aspect of the deep learning paradigm. 

Redundancies of deep learning architectures of various types have been documented in the literature\footnote{For example, vector embeddings are widely used in language models and other dimensionality reduction techniques. Common vector embeddings are invariant under rotational and reflectional symmetries. I will discuss the implications of these symmetries elsewhere and focus here primarily on the permutability of deep networks.}. Unfortunately, they are still poorly understood. It is impossible to fill this gap in one short note, therefore here I focus on one particular redundancy of deep networks due to the permutability of their elements in an attempt to draw attention to the issue.

\subsection{Redundancy of layered structures of permutable elements}

\begin{figure}[ht]
\centering
\includegraphics[width=0.40\linewidth]{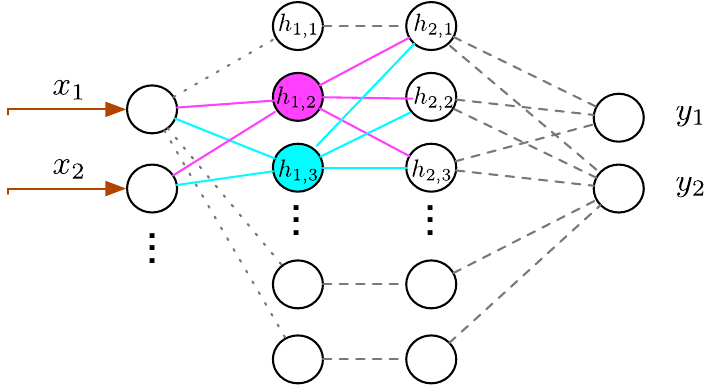}
\includegraphics[width=0.40\linewidth]{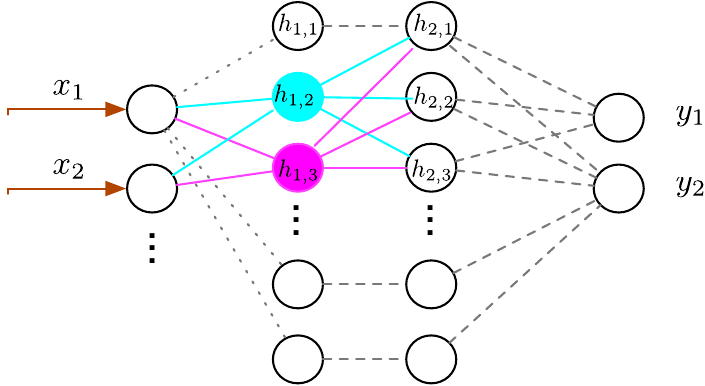} \caption{\label{fig:permutation} Invariance of a two-layer network with respect to the permutation of arbitrary two nodes in a layer. $x$'s are the network's inputs, $y$'s the outputs. Two nodes, $h_{1,2}$ and $h_{1,3}$, of a trained network can be exchanged preserving their inputs and outputs without changing the network's output.}
\end{figure}

The output of a simple two-layer fully connected trained network is invariant with respect to a permutation of nodes within one layer, as illustrated in Fig \ref{fig:permutation} for a toy case of a two-node permutation in a two-layer network.

Following the thorough analysis in \cite{Sorensen_NN_Symm}, I will quickly describe the architecture of a fully connected network of the depth $d$:  
\begin{equation}
    A = (n_0, n_1, \dots, n_{d-1}, n_d) \label{eq:Architecture}
\end{equation}
where $n_i$ is the number of identical elements in the $i$-the layer of the network, $n_0$ is the number of inputs, $n_d$ is the number of outputs. The network is parameterized by
\begin{equation}
    \theta = (\mathbf{W}_1, \dots, \mathbf{W}_d)
\end{equation}
so that $\mathbf{W}_1$ is a $n_0 \times n_1$ matrix, $\mathbf{W}_2$ is a $n_1 \times n_2$ matrix, and so on, to $\mathbf{W}_d$ is a $n_{d-1} \times n_d$ matrix.  

The value of a particular node $i$ in the layer $k$ is then, as usual:
\begin{equation}
    h_{k,j} = \sigma_k\left(\sum_{i=1}^{n_{k-1}} h_{k-1,i} \cdot W_{i,j}(k)\right) \label{eq:hiddeen}
\end{equation}
By the distributive property of the sum in Eq \ref{eq:hiddeen} the output of the function at a node of the layer $k$ does not depend on the order of nodes in the preceding layer $k-1$. Denoting $\pi_i$ a permutation matrix (that is, a matrix containing exactly one 1 in each row and column) in layer $k$, I define the permutation-equivalent network realization as
\begin{equation}
    \pi (\theta )= (\mathbf{W}_1\cdot \pi_1, \pi_1^{-1}\cdot\mathbf{W}_2\cdot\pi_2,\dots,\pi_{d-1}\cdot\mathbf{W}_d])
\end{equation}
Undoubtedly, a permutation (or reordering) of elements in each layer which preserves their connections to the preceding and the following layers does not change the output of the network.

A permutation of many elements qualifies as a strong structural disturbance of the network. A more general result obtained in \cite{NEURIPS2019_04115ec3} is that common network architectures are not "inversely stable", that is, two networks which have similar functionality do not necessarily have similar internal structure   
\begin{equation}
    \|\theta - \theta^\prime\|_\theta < > s\cdot \| \mathcal{F}(\theta) - \mathcal{F}(\theta^\prime)\|_{\mathcal{F}}
\end{equation}
where $\mathcal{F}(\theta)$  is a symbolic representation of the network's functionality. The left and the right sides of this relation define  "closeness" in some appropriate sense for the space of parameterization and the space of the network's functionality, respectively. I will discuss the possible realization of the norm for the parameterization of the network in the section \ref{sec:Remedy_Sorensen}.

The number of possible functionally equivalent structures depends on the deep network architecture (Eq \ref{eq:Architecture}) defined by the number of layers and the number of nodes in each layer. For the case of permutation within each layer it is
\begin{equation}
    \|\pi(A)\| = \prod_{l =1}^{d-1} n_l! \label{eq:numpermutation}
\end{equation}
Eq \ref{eq:numpermutation} implies that one network architecture supports $\|\pi(A)\|$ functionally equivalent structures. This also includes the structure corresponding to an optimal set of parameters $\theta^*$. 

This fact has serious implications for the perspectives of finding a global optimum of the network. The following Corollary I borrow almost verbatim from \cite{Sorensen_NN_Symm} summarizes the permutative redundancy:  
\begin{corollary}
Let $A = (n_0, \dots, n_d) $ be a neural network architecture. Then any optimum of a loss function over the corresponding parameter space has $\|\pi(A)\|$  redundancies - that is, distinct optima that do not correspond to meaningfully distinct parameterizations - in the loss surface. 
\end{corollary}
In other words, the existence of \textit{one} optimum (global or local) immediately implies the existence of $\|\pi(\theta)\|$ equivalent optima in the parameter space.  

The above result is valid for all architectures employing fully connected layers, such as LeNet \cite{LeCunLeNet} and its derivatives \cite{NIPS2012_c399862d}.   For LeNet-5, for example, there are at least  $120!\cdot 80!\cdot 10!$ equivalent global optima, a number greater than the number of atoms in our Universe.

The result applies to other modern connectionist models, such as encoder-decoder and transformer architectures with attention layers over the embedding space. For some architectures, the vector embedding space has other redundancies, such as rotational. I will discuss them elsewhere.

%\centerline{***}
\subsection{Equilibrium stochastic dynamics over multiple redundant optimua}

In section \ref{sec:grad_convergence} I demonstrate that the convergence rate and the convergence outcome critically depend on the shape of the objective in the parameter space. The Hessian is a representation of the objective's local structure around an optimum. To properly assess the implications of the existence of the enormous number of global optima in the optimization space it helps to refer to experimental studies of Hessian eigenvalues in \cite{sagun2017eigenvalueshessiandeeplearning} and \cite{pmlr-v97-ghorbani19b}. 

Authors of \cite{sagun2017eigenvalueshessiandeeplearning} show for toy networks that the bulk of the Hessian spectrum concentrates near zero. Authors argue that discrete positive eigenvalues indicate dependence on the data and the training objective, while the near-zero bulk is a function of the size of the network. Ghorbani et al in \cite{pmlr-v97-ghorbani19b} further develop these ideas and extend their analysis to larger networks.  

Uncertain objective function alone makes the convergence of gradient descent methods problematic. The situation is even more complicated when there are many equivalent global optima in the neighbourhood of each global optimum. 

For a deep network characterized by $d$ layers each $w$ wide, so that $wd=n$, the number of global optima is 
\[
\|\pi\| = (w!)^d \sim \exp \left( n\cdot\log w \right)
\]

On the other hand, the volume occupied by the equilibrium distribution of stochastic gradient iterates is given by Eq \ref{eq:vol_ball}:
\[
V \sim \prod_{i=1}^n \lambda_i^{-1/2} \sim \exp \left(- n/2\cdot\overline{log\lambda} \right)
\]
where 
\[
\overline{log\lambda} = \frac{1}{n}\sum_{i=1}^n \log \lambda_i
\]
The total volume occupied by possible gradient iterates around all global optima is then 
\begin{equation}
    V^* \sim \exp \left( n\cdot \left( \log(w) - \frac{1}{2} \overline{\log\lambda}\right) \right) \label{eq:vstar}
\end{equation}
It grows with the size of the network $n$ at least proportionally to the total volume of the parameter space.  
Next, empirical studies of the Hessian spectrum for deep learning networks (\cite{sagun2017eigenvalueshessiandeeplearning}, \cite{pmlr-v97-ghorbani19b} and references therein) demonstrate that for a wide range of models, from toy ones to large, most of the spectrum is typically concentrated near $\lambda \approx 0$  for untrained models, and $\lambda \gtrsim 0$
for trained models.  The positivity of the Hessian eigenvalues does indicate that the training procedure arrives at a more optimal model. As the size of the network grows, the concentration of eigenvalues near zero seems to increase: 
\begin{equation}
   \frac{ d \overline{\log\lambda}}{d n} > 0
\end{equation}
From Eq \ref{eq:vstar} then it is not unreasonable to make the following 
\begin{conjecture}
For network architecture employing layers of permutable elements, as the network size increases, the stochastic vicinity of each global optimum containing the trajectories of stochastic gradient iterates eventually reaches the vicinities of other global optima. At this point, the approximation of a quadratic, a differentiable, or even a smooth empirical objective function apparently breaks down. The objective landscape becomes a complex web of valleys and ridges covering most of the parameter space, and the gradient iterate trajectories freely traverse between vicinities of multiple equivalent global optima. 
\end{conjecture}

\section{Remedies}

\subsection{Sorensen reordering heuristic} \label{sec:Remedy_Sorensen}
The absence of inverse stability, that is, similarity in functionality does not imply similarity in the internal structure, of standard deep network architectures based on layers of permutationally equivalent elements (see the already mentioned \cite{NEURIPS2019_04115ec3}) is a very inconvenient property.

For mission-critical applications, deep network stability in the course of adaptive re-training is often an important desideratum: not only do stakeholders demand the evidence of the functional continuity of a newly re-trained variant, but the stability and continuity of parameters are required, too, as the evidence of the model self-consistency. That is, if $\theta^*(D)$ is a parameterization of a deep network subject to data $D$, $\mathcal{F}(\theta^*(D))$ is a symbolic representation of its functionality, then it is desired that for small disturbances in the data set, both the deep network functionality and its parameterization must also be small
\begin{equation}
\begin{split}
     \forall D^\prime &: \| D^\prime - D \|_D \ll \|D\|_D \\
              \exists L_{\mathcal{F}}    &: \|\mathcal{F}(\theta^*(D^\prime)) - \mathcal{F}(\theta^*(D)) \|_{\mathcal{F}} < L_{\mathcal{F}}\cdot  \| D^\prime - D \|_D/\|D\|_D \\
               \exists L_{\mathcal{\theta}}   &: \|\theta^*(D^\prime) - \theta^*(D)\|_{\mathcal{\theta}} < L_{\mathcal{\theta}}\cdot  \| D^\prime - D \|_D/\|D\|_D
\end{split} \label{eq:stab_f_theta}
\end{equation}

The substance of $\|\cdot\|_D$, $\|\cdot\|_{\mathcal{F}}$, and $\|\cdot\|_{\mathcal{\theta}}$ is a function of an application. 

Data themselves are easy $\|\cdot\|_D$ might mean cardinality, a set of statistical properties of $D$, Kullback-Leibler divergence of respective distributions, etc.  

The substance of $\|\cdot\|_{\mathcal{F}}$ could be defined via a vector of numeric results of pre-defined regression tests, which fits a typical production deployment requirement, in which case $\|\cdot\|_{\mathcal{F}}$ emphasizes the stability of aggregate functionality, etc. 

The substance of $\|\cdot\|_{\mathcal{\theta}}$ is much more stubborn. Considering the permutative invariance of deep networks and lack of "inverse stability", it is not obvious. Aggregate metrics such as the spectral analysis suggested in \cite{WeightWatchers_2021_Nature} help assess the statistical properties of the network, but do not provide an assessment of the closeness of network parameters or their stability from one re-training to another.

In the rest of this section, I explain how the combination of 
\begin{itemize}
    \item reordering of individual layers and 
    \item the Frobenius normalized metrics
\end{itemize}
provides an approximate assessment of the closeness, in some sense, of two networks of the same architecture. The heuristic has been extensively tested for practical deep learning settings. 

%\subsubsection{Reordering}

The purpose of a reordering algorithm is to order each layer's nodes while preserving the nodes' connectivity. The goal of reordering is to impose a structure onto the network parameterization so that two networks of the same architecture trained with the same data to the same level of optimality have the same or very close internal structures. 

Let $\theta$ be a network configuration and $\theta^\prime$ is a disturbance in the coefficients of its matrices. The disturbance of a matrix is small if some aggregate of the elementwise difference between the disturbed matrix and the original matrix is small. I found a variant of the Frobenius norm works pretty well for practical purposes:
\begin{equation}
    \Phi(\mathbf{W}, \mathbf{W^\prime}))= \frac{\sum_{ij}(W_{ij} - W_{ij}^\prime)^2} {\sum_{ij}(W_{ij} + W_{ij}^\prime)^2}
\end{equation}
If $\mathbf{W}$ and $\mathbf{W^\prime}$ are random matrices of the same structure, $\Phi(\mathbf{W}, \mathbf{W^\prime})) \approx 1$. If they are the same, $\Phi(\mathbf{W}, \mathbf{W^\prime}))= 0$. If $\mathbf{W^\prime}$  is a randomly permuted version of $\mathbf{W}$, their Frobenius norm is close to 1.  The smaller the norm, the greater the correlation between the elements of the two random matrices. Empirically, the above procedure works well for the layers of standard deep architectures. The overall similarity measure between two parameterizations could be defined via a vector of Frobenius norms for all layers.

Due to the immense number of possible permutations (see Eq \ref{eq:numpermutation}), if $\theta^\prime = \pi(\theta)$ is an arbitrary permutation of the nodes in the layers of the network, plus a small disturbance in the network weights, it is not practically possible to recover the original structure of the permuted network to meaningfully compare the two networks by reordering their layers until their Frobenius norm reaches a minimum value.  We thus need a practical reordering algorithm of the rows of $\mathbf{W}(k)$, which provides a fast, albeit likely approximate, solution.    

I list without vouching for any particular one some of the suggested implementations:
\begin{description}
    \item[Lexicographic sorting] Compare the first elements of each row, then the second, etc. This is the ordering heuristic suggested in \cite{Sorensen_NN_Symm}
    \item[Maximin sorting] Find the row $i^*$ in $W_{i^*j}(k)$ for which the spread of its elements is the largest:
    \begin{equation}
        i^* = \argmax_i (\max(W_{i^*j}(k)) - \min(W_{i^*j}(k)) )
    \end{equation}
    Then reorder the columns of $W_(k)$ so that the elements of the $i^*$-th row of $(W_(k)\pi_k$  are sorted.
\end{description}
Both methods with high reliability recover the original structure of a slightly disturbed and permuted matrix. 

Readers can try other ordering heuristics based on treating the elements of $\mathbf{W}(k)$ as random values and comparing various statistics which should preserve the order if the disturbances are small in some sense but can detect large differences between rows.   

The reordering heuristic (sans the Frobenius norm) has been discovered independently by Sorensen in \cite{Sorensen_NN_Symm}  and by the author \cite{VSG2019} who only recently became aware of her published contribution. 

The Sorensen reordering heuristic is not precise and can fail for multiple reasons, such as a wrong selection of the reordering driver $i^*(k)$. Yet, empirically, it works well in many practical applications.

\centerline{***}

The Sorensen reordering requires \[N_S \sim \sum_{k=1}^d n_K\cdot (\log n_k + c)\] operations for a deep network consisting of $d$ layers each of the width $n_k$. It alone does not eliminate the enormous number of ghost local and global minima in traditional deep learning architectures, but perhaps can be a part of a heuristic method to break the permutative symmetry and force the network to adopt and maintain a particular internal structure.

Given the parameterization of the benchmark network of the same architecture, Sorensen reordering can be used at any network re-training stage to track and trace the progress of the network re-training.  

A stopping criterion is another possible use of the Sorensen reordering. For situations of high objective uncertainty, the progress toward optimality within the Boltzmann equilibrium ball \ref{eq:boltzmann} is slow and is not guaranteed to reach the true optimum even if it exists. Therefore, once the retrained network reaches the Boltzmann vicinity of the previously trained network in the parameter space, the gradient iterates can be safely stopped.

I will further explore this line of thought elsewhere.

\subsection{Pre-break the symmetries}

The permutability of interchangeable elements is the source of deep networks' expressive power. Said interchangeability also leads to structural redundancies. Can we preserve deep networks' expressivity by changing their structural elements to reduce or eliminate redundancies?

\subsubsection{Non-permutatiuve pre-pruning}

\label{sec:pre-pruning}
In-training and post-training pruning have already been suggested and used as a regularization and efficiency-increasing method.  I suggest using pre-training pruning with setting some parameters of the network to zero permanently to break the permutative symmetry of network layers and reduce the number of equivalent global optima.  

Permutable connectivity between two consecutive layers $k-1$ and $k$ is expressed via a matrix $\mathbf{W}(k)$ of weights $n_{k-1} \times n_k$  (see Eq \ref{eq:hiddeen}).  

Introduce a binary matrix $\mathbf{B}(k)$ with the following property: 
\begin{equation}
    \forall j,j^\prime : B_{\cdot,j} = b_j \neq  b_{j^\prime} = B_{\cdot,j^\prime} 
\end{equation}
In other words, no two columns of $\mathbf{W}(k)$ are the same. This can easily be achieved with a high probability by populating  $\mathbf{B}(k)$ with random 0s and 1s with a rate $\rho$ of zeros. The probability of a collision remains small for $n_{k-1}, n_k \gg 1$ and can be neglected for practical purposes.

The network layer $k$ which is defined by the pre-pruned matrix of parameters
\begin{equation}
   \tilde{\mathbf{W}}(k) = \mathbf{W}(k) \circ \mathbf{B}(k)
\end{equation}
where $A\circ B$ is the Hadamard product, is non-permutative with a high probability: each connection to the previous layer is almost uniquely encoded by the binary matrix $\mathbf{B}(k)$. 

Removal of the permutativity of layers reduces the astronomical number of equivalent optima but does not guarantee their absence -- e.g. in some degenerate cases of substantially overdetermined networks. 

Lastly, pre-pruning with the rate $\rho$ can reduce the representation power of the network. To avoid it, just to keep the representative power constant, it is reasonable to suggest increasing the number of nodes in the layer by $(1-\rho)^{-1}$ before pre-pruning.

\subsubsection{Ortho-polynomial activations}

The first working example of a deep network was reported by Ivakhnenko and his collaborators who were working in the 1960s and 1970s in what is modern Ukraine ( \cite{IVAKHNENKO1970207}  \cite{Ivakhnenko_1971} and references therein). The authors built a decision-making network based on layers consisting of Kolmogorov-Gabor polynomial activation functions.

In \cite{piterbarg_antonov_2021} the authors revive Ivakhnenko's approach and offer an alternative to deep networks based, again, on families of orthogonal polynomials. The authors find the method particularly suitable for financial applications. 

An orthogonal polynomial basis is an effective nearly universal approximator. The elements of the basis are
\begin{equation}
\begin{split}
    \Phi &= \{ \phi_i(x) \}, i = 1\dots \infty \\
    <\phi_i(x),\phi_j(x)> & = \delta_{ij} \\
    <f(x),g(x)> &= \int_{x_1}^{x_2} f(x)\cdot g(x) \cdot \Psi(x) dx 
\end{split} \label{eq:polynomials}
\end{equation}
where $<\cdot,\cdot>$ denotes the inner product and $\Psi(x)$ and the integration interval $(x_1,x_2)$ are specific for each orthogonal polynomial type. 

By Parceval's theorem for a good enough function $f(x)$
\begin{equation}
    \| f(x) - \sum_{i = 0}^N w_i \cdot \phi_i(x) \|^2 = \|\sum_{i=N+1}^\infty w_i \cdot \phi_i(x) \|^2 < \infty \label{eq:parceval}
\end{equation}
where $\|g(x)\|^2 = <g(x),g(x)>$. Clearly, a polynomial approximator of a degree $N$ cannot represent polynomials of higher degrees, but can be as representative as we need for large enough $N$.  

Basis functions $\phi_i(x)$ represent features of finer scales as $i \rightarrow \infty$.  If the scale of features of $f(x)$ is limited from below by its nature or by a particular application (which is almost always the case), a finite number of basis functions is sufficient to approximate any reasonable function:
\begin{equation}
    \| f(x) - \sum_{i = 0}^N w_i \cdot \phi_i(x) \|^2 < \epsilon_N \|f(x)\|^2 \label{eq:approx_basis} 
\end{equation}

Moreover, an analogue of the Nyquist-Shannon theorem is valid for the orthogonal polynomial approximation: under reasonable assumptions about the behaviour of $f(x)$ and its derivatives, and the scale of features necessary for a particular application, the number of sampling points needed to determine $w_i$ in Eq \ref{eq:approx_basis} cannot be significantly greater than $O(N)$. 

A one-layer Kolmogorov-Gabor-Ivakhnenko polynomial perception-"polytron" of width $N$ is a universal approximation for all functions $f(x)$ for which $\epsilon_N$ in Eq \ref{eq:approx_basis} is small for practical purposes.  

\begin{figure}[ht]
\centering
\includegraphics[width=0.70\linewidth]{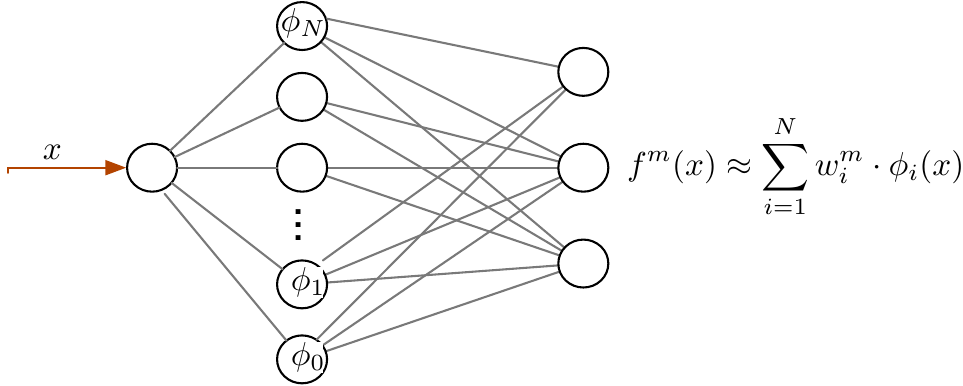}
\caption{\label{fig:polynomial} Polynomial representation of a set of functions $f^m(x)$ as a one-layer network. Binary pre-pruning (see section \ref{sec:pre-pruning} ) ensures non-permutability of the output layer: each $f^m(x)$ represents a different combination of features in the data.} \label{fig:Poly2}. 
\end{figure}

With the polynomial universal approximation the scale of features it captures is easily controlled, various regularizations are interpretable, and the relationship between the data set size necessary to train it seems also easily inferrable from its internal structure.

The gradient iterates corresponding to an output node in  Fig \ref{fig:polynomial} and a quadratic objective function based on the inner product in Eq \ref{eq:polynomials} are:
\begin{equation}
    \nabla_{w_k} J(\mathbf w, x) \sim (f(x) - \sum_{i=0}^N w_i \phi_i(x)) \cdot \Psi(x) \cdot \phi_k(x)
\end{equation}
Evaluated over a batch of data $B = (x,y=f(x))$ the empirical gradient is 
\begin{equation}
    \nabla_{w_k} J(\mathbf w, B) = \sum_x f(x) \cdot \phi_k(x) \cdot \Psi(x) - w_k \cdot \sum_{i=0}^N\phi_i(x) \phi_k(x) \cdot \Psi(x)
\end{equation}

A feasible extension of a simple polynomial representation is a trainable scale parameter at the argument of orthogonal polynomials. If this is the case, the gradient iterates also depend on the derivatives of polynomials.

Polynomial perceptions can be stacked similarly to traditional multi-layer permutable perceptions to form deep and wide networks. Multivariate inputs and internal layers can be handled similarly. Binary pre-pruning I discuss in section \ref{sec:pre-pruning} is also applicable to ensure the non-permutability of stacked polynomial layers. 

Orthogonal polynomials are reasonably computationally efficient. The computation of $\phi_i(x)$ and $d\phi_i(x)/dx$ is often simple recurrent relationships. For example, for the Laguerre polynomials defined on the interval $(0, \infty$, and $\Psi(x) = \exp(-x)$:
\begin{equation}
    \begin{split}
        L_0(x) &= 1 \\
        L_1(x) &= 1- x \\
        L_{i+1}(x) & = \frac{(2i+1-x)L_i(x) - iL_{i-1}(x)}{i+1} \\
        \nabla_x L_i(x) &= i\frac{L_i(x) - L_{i-1}(x)}{x}
    \end{split}
\end{equation}
A polynomial layer output and gradients therefore require $O(N)$ multiplications and additions per next layer's node. 

Practitioners can choose different polynomials for different purposes. For example, standard polynomials defined on a finite range $(-1, 1)$ (or, equivalently, modulo the obvious scaling and translation operation, $(0,1)$) could be particularly suitable for problems where the output of intermediate layers is normalized.  

\subsubsection{Kolmogorov-Arnold Networks (KANs)}

In \cite{liu2024kankolmogorovarnoldnetworks} and \cite{poluektov2024constructionkolmogorovarnoldrepresentationusing} the authors explore the possibility of using arbitrary activation functions parameterized by splines at the nodes of a multilayer network. Network output permits automated differentiation and is as efficient as the deep network consisting of linear permutable elements.  It is a promising approach. I see two non-substantial drawbacks of current KAN architectures which can be easily remedied.
\begin{itemize}
    \item[a)] Splines are less efficient than regular orthogonal polynomials as the basis for representation. Nothing prevents practitioners from replacing the spline-based units with ortho-polynomial units,
    \item[b)] KAN elements, as envisioned by the authors, are still mutually permutable. Therefore, KANs should suffer from the guaranteed existence of multiple equivalent global and local optima which, in turn, guarantees to make the optimization landscape in the case of an uncertain objective a tangled network of ridges and valleys. This limitation does not seem critical: individual nodes, be they splines or polynomial representations, can be easily uniquely constrained and managed via binary pre-pruning. 
\end{itemize}

\subsection{Bio-inspired alternatives to monolithic feedforward architectures and the backpropagation training}

Undeterred by a lack of attention and scarce resources, a few enthusiasts are working on bio-inspired alternatives. I mention just a few. 

Authors of \cite{sanchezgarcia2022efficientmultiscalerepresentationvisual} demonstrate a remarkable result that a bio-inspired visual object recognition system is capable of representing objects in the standard datasets MNIST, F-MNIST, ORL, and CIFAR10 with only 200 neurons, 15 spikes per neuron (a measure of temporal response), and a reasonable set of spatial frequency filters.  
%\begin{wrapfigure}{r}{0.5\textwidth}
\begin{figure}[ht]
\centering
\includegraphics[width=0.60\textwidth]{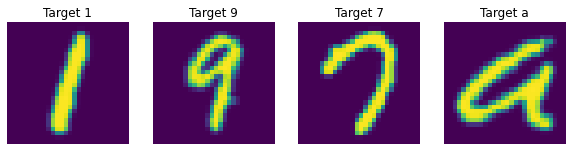} \\
\includegraphics[width=0.60\textwidth]{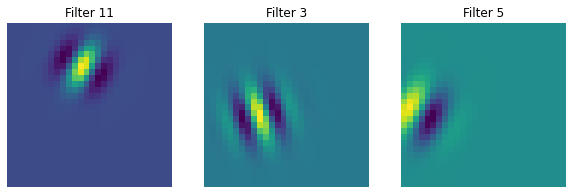} \\
{\includegraphics[width=0.60\textwidth]{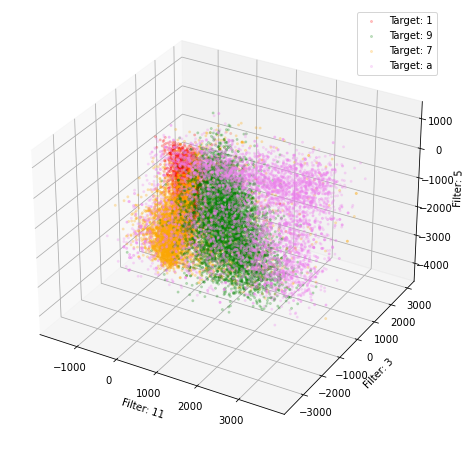}}
 \caption{\label{fig:filters_study}Four EMNIST object targets (top) passed through the three bio-inspired spatial filters (middle), located in the feature space defined by the filter intensities (arbitrary units) (bottom, colour indicates different targets)}    
\end{figure}
%\end{wrapfigure}

The author's own research \cite{VSG2022} confirms this finding from a slightly different perspective. Only 20-30 arbitrary multi-scale filters successfully sort standard EMNIST objects into non-overlapping areas in the feature space (Fig \ref{fig:filters_study}). Once this is achieved, learning a new object is very efficient and incremental, i.e. does not require re-learning from scratch. The recognition system built on these principles turns out to be robust to distortion and noise even when learning does not include distorted and noisy samples. 

State-of-the-art deep learning architectures tackling the same set of problems require hundreds of thousands if not millions of trainable elements. The comparison in efficiency is stunning.  

Authors of \cite{gupta2023bioinspiredlearningbetterbackprop} compare bio-inspired architectures with the standard backpropagation one. They find that bio-inspired architectures learn faster, can use a fraction of data, and outperform traditional backpropagation architectures in accuracy. The so-called direct feedback alignment, one of the approaches used by the authors, is evidently inspired by the existence of multiple feedback pathways in the brains of higher animals. Feedback mechanisms play a crucial role in the processing of visual information in mammals: the non-retinal inputs (i.e. coming not from the eye's retina, but elsewhere in the brain, including the downstream visual apparatus) provide an overwhelming majority of inputs to the lateral geniculate nucleus (LGN) responsible for the early stages of processing of visual stimuli (see \cite{bear2016neuroscience}, \cite{Weyand+2016+135+157} and references therein). 

It is not a secret that a significant part of monolithic deep learning architectures seems to optimize the extraction of basic features from the training dataset. The fact that early layers of trained convolutional networks are essentially spatial filters tuned to the training data set has long been well known. In Krizhevsky's seminal work on image recognition and generation \cite{Krizhevsky2009LearningML} with Restricted Boltzmann Machines, the author has initially achieved very reasonable success with a modular and hierarchical approach resembling the functioning of the mammalian visual system but has ultimately chosen a massive monolithic RBM, a large portion of which was found to be busy learning and using spatial filters. The modular approach was therefore not unpromising. It could have led to more robust, more adaptive, and less expensive visual recognition machines had this line of inquiry been pursued a bit further. 

Another example is large language models. The basic functionality of language models' transformers, at least in some layers, can be understood as simply capturing N-gram rules of the training data set, which can be construed as the most fundamental statistical features of language (see \cite{nguyen2024understandingtransformersngramstatistics} and \cite{svete2024transformersrepresentngramlanguage} and references therein).  

Deep learning architectures are designed to be fine-tuned to a particular environment. They are often brittle and vulnerable to environmental change. Retraining to adapt them to every new environment is necessarily expensive and requires the accumulation of massive amounts of data. What if environments change faster than it takes to accumulate a new data set?  

In contrast, the characteristic of biological systems is their adaptability. It is almost self-evident that a system that is fine-tuned to a particular environment has an increased chance of not surviving even slight environmental changes -- compared with broad-tuned systems. Broadly tuned systems may be less skilled in a particular environment due to the inevitable tradeoffs. But they have a better chance of survival and have a better re-training economics in a changing environment. Broad tuning to many possible classes of environments survives both aleatoric (statistical) and epistemic uncertainty.   

None of the higher functioning biological cognitive machines is monolithic, feedforward, or use backpropagation and gradient optimization. Behind robust, diverse, and adaptable biological systems are two billion years of continuous and successful experimentation. This fact should inspire us.

I will further discuss this line of thought elsewhere.  

\section{Thanks}
The author thanks Vladimir Markov for his helpful comments.

\appendix

\bibliographystyle{alpha}
\bibliography{biblio}

\end{document}